\documentclass[10pt,twocolumn,letterpaper]{article}

\usepackage{iccv}
\usepackage{times}
\usepackage{epsfig}
\usepackage{graphicx}
\usepackage{amsmath}
\usepackage{amssymb}

\usepackage[utf8]{inputenc}
\usepackage{color}
\usepackage{kotex}
\usepackage{cases}
\usepackage{nicefrac}
\usepackage{tabu}
\usepackage{booktabs}
\usepackage{multirow}
\usepackage{subcaption}
\usepackage{caption}
\usepackage{soul}
\usepackage{helvet}  
\usepackage{courier}
\usepackage{url} 
\usepackage{makecell}
\usepackage{enumitem}

\definecolor{orange}{rgb}{1,0.5,0}

\usepackage[pagebackref=true,breaklinks=true,letterpaper=true,colorlinks,bookmarks=false]{hyperref}

\iccvfinalcopy 

\def\iccvPaperID{735} 

\ificcvfinal\fi
\begin{document}

\title{Symmetric Graph Convolutional Autoencoder \\ for Unsupervised Graph Representation Learning}

\vspace{-3cm}
\author{Jiwoong Park$^1$ \quad Minsik Lee$^{2}$ \quad Hyung Jin Chang$^3$ \quad Kyuewang Lee$^{1}$ \quad Jin Young Choi$^1$\\
\vspace{-1mm}
{\small \hspace{1cm}$^1$ASRI, Dept. of ECE., Seoul National University}
{\small \hspace{1cm}$^2$Div. of EE., Hanyang University}\\
\vspace{-1mm}
{\small \hspace{1cm}$^3$School of Computer Science, University of Birmingham} \\
{\tt\scriptsize \{ptywoong,kyuewang,jychoi\}@snu.ac.kr, mleepaper@hanyang.ac.kr, h.j.chang@bham.ac.uk}}

\maketitle

\begin{abstract}
We propose a symmetric graph convolutional autoencoder which produces a low-dimensional latent representation from a graph.
In contrast to the existing graph autoencoders with asymmetric decoder parts, the proposed autoencoder has a newly designed decoder which builds a completely symmetric autoencoder form.
For the reconstruction of node features, the decoder is designed based on Laplacian sharpening as the counterpart of Laplacian smoothing of the encoder, which allows utilizing the graph structure in the whole processes of the proposed autoencoder architecture.
In order to prevent the numerical instability of the network caused by the Laplacian sharpening introduction, we further propose a new numerically stable form of the Laplacian sharpening by incorporating the signed graphs.
In addition, a new cost function which finds a latent representation and a latent affinity matrix simultaneously is devised to boost the performance of image clustering tasks.
The experimental results on clustering, link prediction and visualization tasks strongly support that the proposed model is stable and outperforms various state-of-the-art algorithms.
\end{abstract}

\section{Introduction}
\begin{figure}[t!]
\centering
\begin{subfigure}{.27\textwidth}
\includegraphics[width=\linewidth]{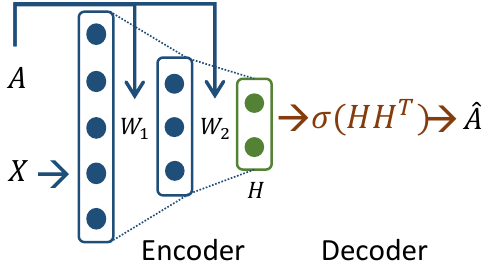}
\caption{VGAE \cite{kipf2016variational}}
\label{VGAE}
\end{subfigure}\hfill
\begin{subfigure}{.2\textwidth}
\includegraphics[width=\linewidth]{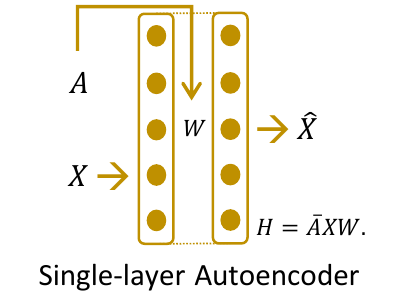}
\caption{MGAE \cite{wang2017mgae}}
\label{MGAE}
\end{subfigure}\par
\begin{subfigure}{.4\textwidth}
\includegraphics[width=\linewidth]{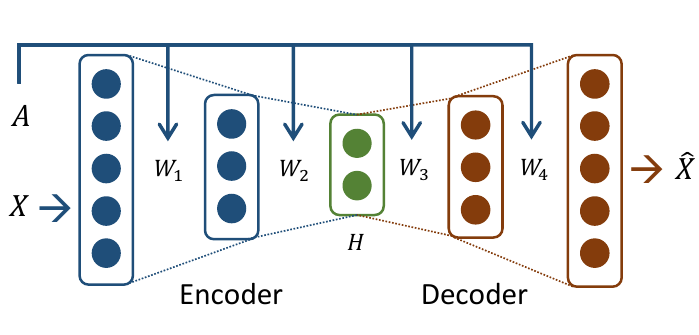}
\caption{Proposed autoencoder}
\label{GALA}
\end{subfigure}
\vspace{-3mm}
\caption{Architectures of existing graph convolutional autoencoders and proposed one. 
$A$, $X$, $H$ and $W$ denote the affinity matrix (structure of graph), node attributes, latent representations and the learnable weight of network respectively.}
\label{comparison}
\end{figure}

A graph, which consists of a set of nodes and edges, is a powerful tool to seek the geometric structure of data.
There are various applications using graphs in the machine learning and data mining fields such as node clustering \cite{ng2002spectral}, dimensionality reduction \cite{belkin2002laplacian}, social network analysis \cite{lazer2009life}, chemical property prediction of a molecular graph \cite{duvenaud2015convolutional}, and image segmentation \cite{shi2000normalized}. 
However, conventional methods for analyzing a graph have several problems such as low computational efficiency due to eigendecomposition or singular value decomposition, or only showing a shallow relationship between nodes.

In recent years, an emerging field called \textit{geometric deep learning} \cite{bronstein2017geometric}, generalizes deep neural network models to non-Euclidean domains such as meshes, manifolds, and graphs 
\cite{kipf2017semi,monti2017geometric,litany2017deformable}.
Among them, finding deep latent representations of geometrical structures of graphs using an autoencoder framework is getting growing attention.
The first attempt is VGAE \cite{kipf2016variational} which consists of a Graph Convolutional Network (GCN) \cite{kipf2017semi} encoder and a matrix outer-product decoder as shown in Figure \ref{comparison} (a). 
As a variant of VGAE, ARVGA \cite{pan2018adversarially} has been proposed by incorporating an adversarial approach to VGAE.
However, VGAE and ARVGA were designed to reconstruct the affinity matrix $A$ instead of node feature matrix $X$.
Hence, the decoder part cannot be learnable, therefore, 
the graphical feature cannot be used at all in the decoder part. 
These facts can degrade the capability of graph learning.
Following that, MGAE \cite{wang2017mgae} has been proposed, which uses stacked single layer graph autoencoder with linear activation function and marginalization process as shown in Figure \ref{comparison} (b).
However, since the MGAE reconstructs the feature matrix of nodes without hidden layers,
it cannot manipulate the dimension of the latent representation and performs a linear mapping.
This is a distinct limitation in finding a latent representation that clearly reveals the structure of the graph.

To overcome the limitation of the existing 
graph convolutional autoencoders, in this paper, we propose a novel graph convolutional autoencoder framework which has symmetric autoencoder architecture and uses both graph and node attributes in both the encoding and decoding processes as illustrated in Figure \ref{comparison} (c).
Our design of the decoder part is motivated from the analysis in a recent paper \cite{li2018deeper}, that the encoder of 
VGAE \cite{kipf2016variational}
can be interpreted as a special form of Laplacian smoothing \cite{taubin1995signal} that computes the new representation of each node as a weighted local average of neighbors and itself. 
This interpretation has inspired us to design 
a decoder to perform \emph{Laplacian sharpening}, which is a counterpart of Laplacian smoothing.
To realize a decoder to do Laplacian sharpening, we express Laplacian sharpening in the form of Chebyshev polynomial and newly reformulate it in a numerically stable form by utilizing a signed graph \cite{li2009note}. 

In computer vision fields, there is a popular assumption that, even though image datasets are high-dimensional in their ambient spaces, they usually reside in multiple low-dimensional subspaces \cite{vidal2011subspace}. 
Thus, especially for image clustering tasks, we apply the concept of subspace clustering, which has such an assumption about the input data in its own definition, to our graph convolutional autoencoder framework.
Specifically, to find a latent representation and a latent affinity matrix simultaneously, we merge a subspace clustering cost function into the reconstruction cost of the autoencoder.
Contrary to the conventional subspace clustering cost function \cite{ji2017deep, zhou2018deep}, we could derive a computationally efficient cost function.

The main contributions of this paper are summarized as follows: 
\begin{itemize}[leftmargin=*]
\item We propose the first completely symmetric graph convolutional autoencoder which utilizes both the structure of the graph and node attributes through the whole encoding-decoding process.
\item We derive a new numerically stable form of decoder preventing the numerical instability of the neural network.
\item We design a computationally efficient subspace clustering cost to find both latent representation and a latent affinity matrix simultaneously for image clustering tasks.
\end{itemize}

In experiments, the validity of the proposed components is shown by doing ablation experiments on our architecture and cost function. 
Also, the superior performance of the proposed method is validated by comparing it with the state-of-the-art methods and visualizing the graph clustered by our framework.
\section{Preliminaries}

\label{section2}
\subsection{Basic notations on graphs}
A graph is represented as $\mathcal{G}=(\mathcal{V}, \mathcal{E}, A)$, where $\mathcal{V}$ 
denotes the node set with $v_i \in \mathcal{V}$ and $|\mathcal{V}| = n$, $\mathcal{E}$ denotes the edge set with $(v_i, v_j) \in \mathcal{E}$, and $A \in {\rm I\!R}^{n\times n}$ denotes an affinity matrix which encodes pairwise affinities between nodes. 
$D \in {\rm I\!R}^{n\times n}$ denotes a degree matrix which is a diagonal matrix with 
$D_{ii} = \sum_{j}A_{ij}$. 
Unnormalized graph Laplacian $\Delta$ is defined by $\Delta = D - A$ \cite{chung1997spectral}. Symmetric graph Laplacian $L$ and random walk graph Laplacian $L_{rw}$ are defined by $L = I_n - D^{-\frac{1}{2}}AD^{-\frac{1}{2}}$ and $L_{rw} = I_n - D^{-1}A$ respectively, where $I_n \in {\rm I\!R}^{n\times n}$ denotes an identity matrix. 
Note that the $\Delta$, $L$ and $L_{rw}$ are positive semidefinite matrices.
\subsection{Spectral convolution on graphs}
A spectral convolution on a graph \cite{shuman2013emerging} is the multiplication 
of an input signal $x \in {\rm I\!R}^{n}$ with 
a spectral filter $g_{\theta} = diag(\theta)$ parameterized by the vector of Fourier coefficients $\theta \in {\rm I\!R}^{n}$ as follows:
\begin{equation}
g_{\theta} * x = U g_{\theta} U^{T} x,
\end{equation}
where $U$ 
is the matrix of eigenvectors 
of the symmetric graph Laplacian $L = U \Lambda U^{T}$.
$U^{T} x$ is the graph Fourier transform of the input $x$, and $g_{\theta}$ is a function of the eigenvalues of $L$, i.e., $g_{\theta}(\Lambda)$, where $\Lambda$ is the diagonal matrix of eigenvalues of $L$.
However, this operation is inappropriate for large-scale graphs since it requires an eigendecomposition to obtain the eigenvalues and eigenvectors of $L$.
To avoid computationally expensive operations, the spectral filter $g_{\theta} (\Lambda)$ was approximated by $K^{th}$ order Chebyshev polynomials in previous works \cite{hammond2011wavelets}.
By doing so, the spectral convolution on the graph can be approximated as
\begin{equation}
g_{\theta} * x \approx U \sum_{k = 0}^{K} \theta_k' T_k(\tilde{\Lambda}) U^{T} x = \sum_{k = 0}^{K} \theta_k' T_k(\tilde{L})x,
\end{equation}
where $T_k(\cdot)$ and $\theta'$ denote the Chebyshev polynomials and a vector of the Chebyshev coefficients respectively. 
$\tilde{\Lambda}$ is $\frac{2}{\lambda_{max}}\Lambda - I_n$, $\lambda_{max}$ denotes the largest eigenvalue of $L$ and $\tilde{L}$ is $U \tilde{\Lambda} U^{T} = \frac{2}{\lambda_{max}}L - I_n$.
The approximated model above
is used as a building block of a convolution
on graphs in \cite{defferrard2016convolutional}.

In the GCN \cite{kipf2017semi}, the Chebyshev approximation model was simplified by setting $K = 1$, $\lambda_{max} \approx 2$ and $\theta = \theta_0' = -\theta_1'$. 
This makes the spectral convolution simplified as follows:
\begin{equation}
g_{\theta} * x \approx \theta(I_n + D^{-\frac{1}{2}}AD^{-\frac{1}{2}})x.
\label{simplified_conv}
\end{equation}
However, repeated application of $I_n + D^{-\frac{1}{2}}AD^{-\frac{1}{2}}$ can cause 
numerical instabilities
in neural networks since the spectral radius of $I_n + D^{-\frac{1}{2}}AD^{-\frac{1}{2}}$ is $2$, and the Chebyshev polynomials form an orthonormal basis when its spectral radius is $1$.
To circumvent this issue, the GCN uses renormalization trick:
\begin{equation}
I_n + D^{-\frac{1}{2}}AD^{-\frac{1}{2}} \rightarrow \tilde{D}^{-\frac{1}{2}} \tilde{A} \tilde{D}^{-\frac{1}{2}},
\end{equation}
where $\tilde{A} = A + I_n$ and $\tilde{D}_{ii} = \sum_{j} \tilde{A}_{ij}$.
Since adding self-loop on nodes to an affinity matrix cannot affect the spectral radius of the corresponding graph Laplacian matrix \cite{grigorchuk1999asymptotic}, this renormalization trick can provide a numerically stable form of $I_n + D^{-\frac{1}{2}}AD^{-\frac{1}{2}}$ while maintaining the meaning of each elements as follows:
\begin{equation}
(I_n + D^{-\frac{1}{2}}AD^{-\frac{1}{2}})_{ij}=\begin{cases}
   1& i = j
   \\
   A_{ij}/\sqrt{D_{ii} D_{jj}}& i \neq j
\end{cases}
\end{equation}
\begin{equation}
(\tilde{D}^{-\frac{1}{2}}\tilde{A}\tilde{D}^{-\frac{1}{2}})_{ij}=\begin{cases}
   1/(D_{ii} + 1)& i = j
   \\
   A_{ij}/\sqrt{(D_{ii}+1)(D_{jj}+1)}& i \neq j.
\end{cases}
\end{equation}
Finally, 
the forward-path of the GCN can be expressed by
\begin{equation}
H^{(m+1)} = \xi(\tilde{D}^{-\frac{1}{2}} \tilde{A} \tilde{D}^{-\frac{1}{2}} H^{(m)} \Theta^{(m)}),
\label{stable_L_smoothing}
\end{equation}
where $H^{(m)}$ is the activation matrix in the $m^{th}$ layer and $H^{(0)}$ is the input nodes' feature matrix $X$. 
$\xi(\cdot)$ is a nonlinear activation function like ReLU($\cdot$) = $\max(0, \cdot)$, and $\Theta^{(m)}$ is a trainable weight matrix. 
The GCN presents a computationally efficient convolutional process (given the assumption that $\tilde{A}$ is sparse) and achieves an improved accuracy over the state-of-the-art methods
in semi-supervised node classification task by using features of nodes and geometric structure of graph simultaneously.

\subsection{Laplacian smoothing}
Li et al. \cite{li2018deeper} demystify GCN \cite{kipf2017semi} and show that GCN is a special form of Laplacian smoothing \cite{taubin1995signal}. 
Laplacian smoothing is a process that calculates a new representation of the input as a weighted local average of its neighbors and itself. 
When we add a self-loop on the nodes, the affinity matrix becomes $\tilde{A} = A + I_n$ and the degree matrix becomes $\tilde{D} = D + I_n$.
Then, the Laplacian smoothing equation is given as follows:
\begin{equation}
x_i^{(m+1)} = (1 - \gamma)x_i^{(m)} + \gamma \sum_{j} \frac{\tilde{A}_{ij}}{\tilde{D}_{ii}} x_j^{(m)},
\end{equation}
where $x_i^{(m+1)}$ is the new representation of $x_i^{(m)}$, and $\gamma$ $(0 < \gamma \leq 1)$ is a regularization parameter which controls the importance between itself and its neighbors.
We can rewrite the above equation in a matrix form as follows:
\begin{align}
X^{(m+1)} &= (1 - \gamma)X^{(m)} + \gamma \tilde{D}^{-1} \tilde{A} X^{(m)}\nonumber\\
          &= X^{(m)} - \gamma (I_n - \tilde{D}^{-1} \tilde{A}) X^{(m)} \label{original_smoothing} \\
          &= X^{(m)} - \gamma \tilde{L}_{rw} X^{(m)}\nonumber.
\end{align}
If we set $\gamma = 1$ and replace $\tilde{L}_{rw}$ with $\tilde{L}$, then Eq. (\ref{original_smoothing}) is changed into $X^{(m+1)} = \tilde{D}^{-\frac{1}{2}} \tilde{A} \tilde{D}^{-\frac{1}{2}}X^{(m)}$ and this equation is the same as the renormalized version of spectral convolution in Eq. (\ref{stable_L_smoothing}).
From the above interpretation, Li et al. explain that the superior performance of GCN in semi-supervised node classification task is due to Laplacian smoothing which makes the features of nodes in the same clusters become similar.

\section{The proposed method}
\label{section3}
In this section, we propose a novel graph convolutional autoencoder framework, named as GALA (Graph convolutional Autoencoder using LAplacian smoothing and sharpening). 
In GALA, there are $M$ layers in total, from the first to $\tfrac{M}{2}$th layers for the encoder and from the $\left(\tfrac{M}{2}+1\right)$th to $M$th layers for the decoder where $M$ is an even number.
The encoder part of GALA is designed to perform the computationally efficient spectral convolution on the graph with a numerically stable form of Laplacian smoothing in the Eq. (\ref{stable_L_smoothing}) \cite{kipf2017semi}.
Along with this, its decoder part is designed to be a special form of Laplacian sharpening \cite{taubin1995signal}, unlike the existing VGAE-related algorithms. 
By this decoder part, GALA reconstructs the feature matrix of nodes directly, instead of yielding an affinity matrix as in the existing VGAE-related algorithms whose decoder parts are incomplete.
Furthermore, to enhance the performance of image clustering, we devise a computationally efficient subspace clustering cost term which is added to the reconstruction cost of GALA.

\subsection{Laplacian sharpening}
Because the encoder performs Laplacian smoothing that makes the latent representation of each node similar to those of its neighboring nodes,  
we design the decoder part to perform Laplacian sharpening as the counterpart of Laplacian smoothing. 
Laplacian sharpening is a process that makes the reconstructed feature of each node farther away from the centroid of its neighbors, which accelerates the reconstruction along with the reconstruction cost and is governed by
\begin{equation}
x_i^{(m+1)} = (1 + \gamma)x_i^{(m)} - \gamma \sum_{j} \frac{A_{ij}}{D_{ii}} x_j^{(m)},
\label{original_sharpening}
\end{equation}
where $x_i^{(m+1)}$ is the new representation of $x_i^{(m)}$, and $\gamma$ is the regularization parameter which controls the importance between itself and its neighbors.
The matrix form of Eq. (\ref{original_sharpening}) is given by 
\begin{align}
X^{(m+1)} &= (1 + \gamma)X^{(m)} - \gamma D^{-1} A X^{(m)}\nonumber\\
          &= X^{(m)} + \gamma (I_n - D^{-1} A) X^{(m)} \\
          &= X^{(m)} + \gamma L_{rw} X^{(m)}\nonumber.
\end{align}
Analogous to the encoder, we set $\gamma = 1$ and replace $L_{rw}$ with $L$.
Similar to Eq. (\ref{simplified_conv}),
we can express Laplacian sharpening in the form of Chebyshev polynomial and simplify it with $K = 1$, $\lambda_{max} \approx 2$, and $\theta = \frac{1}{2} \theta_0 ' = \theta_1 '$. 
Then, a decoder layer can be expressed by 
\begin{equation}
H^{(m+1)} = \xi((2I_n - D^{-\frac{1}{2}} A D^{-\frac{1}{2}}) H^{(m)} \Theta^{(m)}),
\label{propagation}
\end{equation}
where $H^{(m)}$ is the matrix of the activation in the $m^{th}$ layer, $2I_n - D^{-\frac{1}{2}} A D^{-\frac{1}{2}}$ is a special form of Laplacian sharpening, $\xi(\cdot)$ is the nonlinear activation function like ReLU($\cdot$) = $\max(0, \cdot)$, and $\Theta^{(m)}$ is a trainable weight matrix.
However, since the spectral radius of $2I_n - D^{-\frac{1}{2}} A D^{-\frac{1}{2}}$ is $3$, repeated application of this operator can be numerically instable.
Hence, as GCN finds a numerically stable form of Chebyshev polynomials,
we have to find a numerically stable form of Laplacian sharpening while maintaining its meaning.

\subsection{Numerically stable Laplacian sharpening}
To find a new representation of Laplacian sharpening whose spectral radius is 1, we use a signed graph \cite{li2009note}.
A signed graph is denoted by $\Gamma=(\mathcal{V}, \mathcal{E}, \hat{A})$ which is induced from the unsigned graph $\mathcal{G}=(\mathcal{V}, \mathcal{E}, A)$, where each element in $\hat{A}$ has the same absolute value with $A$, but its sign is changed into minus or keeps plus.
The degree matrix of the signed graph $\Gamma$ is denoted by $\hat{D}$ which is obtained from $\hat{A}$. 
In the signed graph, a problem occurs when calculating the degree matrix $\hat{D}$ by the conventional way that may cancel the mixed signed weights in summation and so fails to yield the degree value representing the connectivity of a node to its neighbors.
Thus, by following the practice for signed graphs, we calculate the degree of each node by $\hat{D}_{ii} = \sum_{j} |\hat{A}_{ij}|$ that has the same value (degree of connectivity) as in the unsigned graph.
By using $\hat{A}$ and $\hat{D}$, we can construct an unnormalized graph Laplacian $\hat{\Delta} = \hat{D} - \hat{A}$ and symmetric graph Laplacian $\hat{L} = I_n - \hat{D}^{-\frac{1}{2}}\hat{A}\hat{D}^{-\frac{1}{2}}$ of the signed graph.
From Theorem 1 of \cite{li2009note}, the range of the eigenvalue of $\hat{L}$ is $[0,2]$, thus the spectral radius of $\hat{D}^{-\frac{1}{2}}\hat{A}\hat{D}^{-\frac{1}{2}}$ is 1 for any choice of $\hat{A}$.
Using this result, instead of Eq. (\ref{propagation}), we use a numerically stable form of Laplacian sharpening with spectral radius of 1, given by
\begin{equation}
H^{(m+1)} = \xi(\hat{D}^{-\frac{1}{2}}\hat{A}\hat{D}^{-\frac{1}{2}} H^{(m)} \Theta^{(m)}).
\label{propagation_stable}
\end{equation}
The remaining issue is  
to choose $\hat{A}$ induced from $A$ so that $\hat{D}^{-\frac{1}{2}}\hat{A}\hat{D}^{-\frac{1}{2}}$ maintains the meaning of each element of $2I_n - D^{-\frac{1}{2}} A D^{-\frac{1}{2}}$ in Eq. (\ref{propagation}). To achieve this, we 
map all weights of the unsigned $A$ to negative weights and adding a self-loop 
with a weight value $2$ to each node, that is, 
$\hat{A} = 2I_n - A$ and $\hat{D} = 2I_n + D$.
Then, each element 
of $\hat{D}^{-\frac{1}{2}}\hat{A}\hat{D}^{-\frac{1}{2}}$ is obtained by 
\begin{equation}
(\hat{D}^{-\frac{1}{2}}\hat{A}\hat{D}^{-\frac{1}{2}})_{ij}=\begin{cases}
   2/(D_{ii} + 2)& i = j
   \\
   -A_{ij}/\sqrt{(D_{ii}+2)(D_{jj}+2)}& i \neq j,
\end{cases}
\label{stable_L_sharpening}
\end{equation}
which has the same meaning with the original one given by
\begin{equation}
(2I_n - D^{-\frac{1}{2}} A D^{-\frac{1}{2}})_{ij}=\begin{cases}
   2& i = j
   \\
   -A_{ij}/\sqrt{D_{ii} D_{jj}}& i \neq j.
\end{cases}
\label{L_sharpening}
\end{equation}

\begin{figure*}[t!]
\centering
\begin{subfigure}{.33\textwidth}
\centering
\includegraphics[height=1.3cm, trim = {0.5cm 1.8cm 1.1cm 1.4cm}, clip]{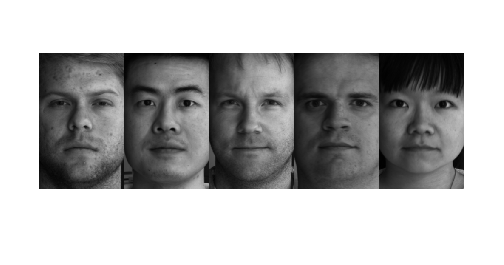}
\caption{YALE}
\end{subfigure}\hfill
\begin{subfigure}{.33\textwidth}
\centering
\includegraphics[height=1.3cm, trim = {2.8cm 1.6cm 6cm 1cm}, clip]{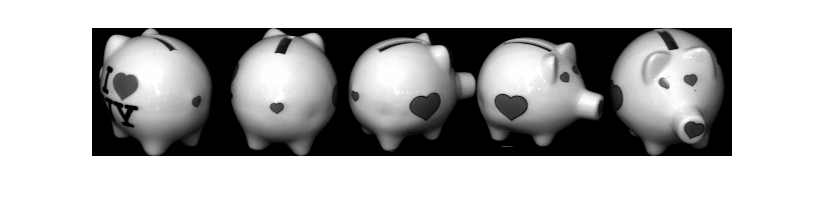}
\caption{COIL20}
\end{subfigure}\hfill
\begin{subfigure}{.33\textwidth}
\centering
\includegraphics[height=1.3cm, trim = {2.5cm 2.2cm 2.5cm 1cm}, clip]{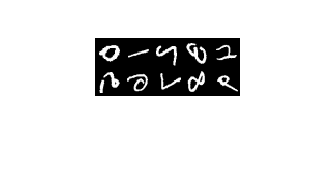}
\caption{MNIST}
\end{subfigure}\hfill
\vspace{-3mm}
\caption{
Sample images of three image datasets
}
\vspace{-2mm}
\label{sample_image}
\end{figure*}
From Eqs. (\ref{propagation_stable}), (\ref{stable_L_sharpening}) and (\ref{L_sharpening}), the numerically stable decoder layer of GALA is given as
{\small
\begin{align}
H^{(m+1)} = \xi(\hat{D}^{-\frac{1}{2}}\hat{A}\hat{D}^{-\frac{1}{2}}  H^{(m)}\Theta^{(m)}), (m = \tfrac{M}{2},...,M-1),
\label{prop_enc}
\end{align}
} 
where $\hat{A} = 2I_n - A$ and $\hat{D} = 2I_n + D$.
The encoder part of GALA is constructed by using Eq. (\ref{stable_L_smoothing}) as in GCN \cite{kipf2017semi} as  
{\small
\begin{align}
H^{(m+1)} = \xi(\tilde{D}^{-\frac{1}{2}} \tilde{A} \tilde{D}^{-\frac{1}{2}} H^{(m)} \Theta^{(m)}), (m = 0,...,\tfrac{M}{2}-1), 
\label{prop_dec}
\end{align}
} 
where $H^{(0)}=X$ is the feature matrix of the input nodes, $\tilde{A} = I_n + A$ and $\tilde{D} = I_n + D$.
The complexity of propagation functions, Eqs. (\ref{prop_enc}) and (\ref{prop_dec}), are both $\mathcal{O}(mpc)$, where $m$ is the cardinality of edges in the graph, $p$ is the feature dimension of the previous layer, and $c$ is the feature dimension of the current layer.
\begin{table}[t]
\caption{Effectiveness of various decoders}
\vspace{-3mm}
\footnotesize
\begin{tabu}{c | X[c] X[c] X[c] | X[c] X[c] X[c]}
\toprule[1pt]
\multirow{2}{*}{} & & Cora & & & Citeseer & \\
\cmidrule{2-7}
 & ACC & NMI & ARI & ACC & NMI & ARI \\
\midrule
Eq. (\ref{stable_L_smoothing}) & 0.5628 & 0.4074 & 0.3289 & 0.5296 & 0.2588 & 0.2437 \\
Eq. (\ref{propagation}) & 0.5999 & 0.4274 & 0.3775 & 0.5915 & 0.3177 & 0.3126\\
Eq. (\ref{prop_enc}) & \textbf{0.7459} & \textbf{0.5767} & \textbf{0.5315} & \textbf{0.6932} & \textbf{0.4411} & \textbf{0.4460} \\
\bottomrule[1pt]
\end{tabu}
\vspace{-2mm}
\label{decoders}
\end{table}
Since the complexity is linear in the number of edges in the graph, the proposed algorithm is computationally efficient (given the assumption that $A$ is sparse).
Also, from Eq. (\ref{prop_dec}), since GALA decodes the latent representation using both the graph structure and node features, the enhanced decoder of GALA can help to find more distinct latent representation.

In Table \ref{decoders}, we show the reason why the Laplacian smoothing is not appropriate to the decoder and the necessity of numerically stable Laplacian sharpening by node clustering experiments (the higher values imply the more correct results).
Laplacian smoothing decoder (Eq. \ref{stable_L_smoothing}) shows the lowest performances, since Laplacian smoothing which makes the representation of each node similar to those of its neighboring nodes conflicts with the purpose of reconstruction cost.
A numerically instable form of Laplacian sharpening decoder (Eq. \ref{propagation}) shows higher performance compared to smoothing decoder because the role of Laplacian sharpening coincide with reconstructing the node feature.
The performance of proposed numerically stable Laplacian sharpening decoder (Eq. \ref{prop_enc}) significantly higher than others, since it solves instability issue of neural network while maintaining the meaning of original Laplacian sharpening.

The basic cost function of GALA is given by
\begin{equation}
\begin{aligned}
& \underset{\bar{X}}{\text{min}}
& & \frac{1}{2}{\| X - \bar{X} \|}^2_F, 
\end{aligned}
\label{recon_loss}
\end{equation}
where $\bar{X}$ is the reconstructed feature matrix of nodes, 
the column of $\bar{X}$ corresponds to the output of the decoder for an input feature of a node, and ${\| \cdot \|}_F$ denotes the Frobenius norm.

\subsection{Subspace clustering cost for image clustering}
It is a well-known assumption that image datasets are often drawn from multiple low-dimensional subspaces, although their data dimensions are high. 
Accordingly, subspace clustering, which has such an assumption about the input data in its own definition, has shown prominent clustering performance on various image datasets.
Hence, we add an element of subspace clustering to the proposed method in the case of image clustering tasks.
Among the various subspace clustering models, we add Least Squares Regression (LSR) \cite{lu2012robust} model for computational efficiency.
Then the cost function for training of GALA becomes
\begin{equation}
{\small
\begin{aligned}
& \underset{\bar{X}, H, A_H}{\text{min}}
& & \frac{1}{2}{\| X - \bar{X} \|}^2_F + \frac{\lambda}{2}{\| H - HA_H \|}^2_F + \frac{\mu}{2}{\| A_H \|}^2_F, 
\end{aligned}}
\label{LSR_loss}
\end{equation}
where $H \in {\rm I\!R}^{k\times n}$ denotes the latent representations (i.e., the output of the encoder), $A_H \in {\rm I\!R}^{n\times n}$ denotes the affinity matrix which is a new latent variable for subspace clustering, and $\lambda, \mu$ are the regularization parameters.
The second term of Eq. (\ref{LSR_loss}) aims at the self-expressive model of subspace clustering and the third term of Eq. (\ref{LSR_loss}) is for regularizing $A_H$.
If we only consider minimizing $A_H$, the problem becomes:
\begin{equation}
\begin{aligned}
& \underset{A_H}{\text{min}}
& & \frac{\lambda}{2}{\| H - HA_H \|}^2_F + \frac{\mu}{2}{\| A_H \|}^2_F.
\end{aligned}
\end{equation}
We can easily obtain the analytic solution $A_H^{*} = (H^{T}H + \frac{\mu}{\lambda}I_n)^{-1}H^{T}H$ by the fact that LSR model is quadratic on the variable $A_H$.
By using this analytic solution and singular value decomposition, we derive a computationally efficient subspace clustering cost function as follows (The details are reported in the supplementary material):
\begin{equation}
\begin{aligned}
& \underset{\bar{X}, H}{\text{min}}
& & \frac{1}{2}{\| X - \bar{X} \|}^2_F + \frac{\mu \lambda}{2}tr((\mu I_k + \lambda HH^{T})^{-1} HH^{T}), 
\end{aligned}
\label{final_loss}
\end{equation}
where $tr( \cdot )$ denotes the trace of the matrix.
The above problem can be solved by $k\times k$ matrix inversion instead of $n\times n$ matrix inversion. 
Since the dimension of latent representation ($k$) is much smaller than the number of nodes ($n$), this simplification can reduce the computational burden significantly 
from $\mathcal{O}(n^3)$ to $\mathcal{O}(k^3)$.
\subsection{Training}
We train GALA to minimize Eq. (\ref{recon_loss}) by using the ADAM algorithm \cite{kingma2014adam}.
We train GALA deterministically by using the full batch in each training epoch and stop when the cost is converged, so the number of epochs of each dataset varies. 
Note here that using the full batch during training is a common approach in neural networks based on spectral convolution on graph.
Specifically, we set the learning rate to $1.0 \times 10^{-4}$ for training and train GALA in an unsupervised way without any cluster labels.
When the subspace clustering cost is added to reconstruction cost for image clustering tasks, we use pre-training and fine-tuning strategies similar to the ones in \cite{ji2017deep} to train GALA.
First, in the pre-training stage, the training method is the same as that of minimizing Eq. (\ref{recon_loss}).
After pre-training, we fine-tune GALA to minimize Eq. (\ref{final_loss}) using ADAM.
As in the pre-training, we train GALA deterministically by using full batch in each training epoch, and we set the number of epochs of the fine-tuning stage as 50 for all dataset.
We set the learning rate to $1.0 \times 10^{-6}$ for fine-tuning.

After the training process are over, we construct $k$-nearest neighborhood graph using attained latent representations $H^*$.
Then we perform spectral clustering \cite{ng2002spectral} and get the clustering performance.
In the case of image clustering, after all training processes are over, we construct the optimal affinity matrix $A_H^*$ noted in the previous subsection by using the attained latent representation matrix $H^*$ from GALA.
Then we perform spectral clustering \cite{ng2002spectral} on the affinity matrix and get the optimal clustering with respect to our cost function. 
\begin{table*}[h]
\caption{Experimental results of node clustering}
\vspace{-3mm}
\footnotesize
\begin{tabu}{c | X[c] X[c] X[c] | X[c] X[c] X[c] | X[c] X[c] X[c]}
\toprule[1pt]
 \multirow{2}{*}{} & & Cora & & & Citeseer & & & Wiki & \\
 \cmidrule{2-10}
 & ACC & NMI & ARI & ACC & NMI & ARI & ACC & NMI & ARI  \\
\midrule
\midrule
 Kmeans\cite{lloyd1982least} & 0.4922 & 0.3210 & 0.2296& 0.5401 & 0.3054 & 0.2786& 0.4172 & 0.4402 & 0.1507\\
 \midrule
 Spectral\cite{ng2002spectral} & 0.3672 & 0.1267 & 0.0311& 0.2389 & 0.0557 & 0.0100& 0.2204 & 0.1817 & 0.0146\\
 Big-Clam\cite{yang2013overlapping} & 0.2718 & 0.0073 & 0.0011& 0.2500 & 0.0357 & 0.0071& 0.1563 & 0.0900 & 0.0070\\
 DeepWalk\cite{perozzi2014deepwalk} & 0.4840 & 0.3270 & 0.2427& 0.3365 & 0.0878 & 0.0922& 0.3846 & 0.3238 & 0.1703 \\
 GraEnc\cite{tian2014learning} & 0.3249 & 0.1093 & 0.0055& 0.2252 & 0.0330 & 0.0100& 0.2067 & 0.1207 & 0.0049\\
 DNGR\cite{cao2016deep} & 0.4191 & 0.3184 & 0.1422& 0.3259 & 0.1802 & 0.0429& 0.3758 & 0.3585 & 0.1797 \\
 \midrule
 Circles\cite{leskovec2012learning} & 0.6067 & 0.4042 & 0.3620& 0.5716 & 0.3007 & 0.2930& 0.4241 & 0.4180 & 0.2420 \\
 RTM\cite{chang2009relational} & 0.4396 & 0.2301 & 0.1691& 0.4509 & 0.2393 & 0.2026& 0.4364 & 0.4495 & 0.1384 \\
 RMSC\cite{xia2014robust} & 0.4066 & 0.2551 & 0.0895& 0.2950 & 0.1387 & 0.0488& 0.3976 & 0.4150 & 0.1116 \\
 TADW\cite{yang2015network} & 0.5603 & 0.4411 & 0.3320& 0.4548 & 0.2914 & 0.2281& 0.3096 & 0.2713 & 0.0454\\ 
 \midrule
 VGAE\cite{kipf2016variational} & 0.5020 & 0.3292 & 0.2547& 0.4670 & 0.2605 & 0.2056& 0.4509 & 0.4676 & 0.2634\\
 MGAE\cite{wang2017mgae} & 0.6844 & 0.5111 & 0.4447& 0.6607 & 0.4122 & 0.4137 & 0.5146 & 0.4852 & 0.3490 \\
 ARGA\cite{pan2018adversarially} & 0.6400 & 0.4490 & 0.3520& 0.5730 & 0.3500 & 0.3410 & 0.3805 & 0.3445 & 0.1122 \\
 ARVGA\cite{pan2018adversarially} & 0.6380 & 0.4500 & 0.3740& 0.5440 & 0.2610 & 0.2450 & 0.3867 & 0.3388 & 0.1069 \\ 
 \midrule
 GALA & \textbf{0.7459} & \textbf{0.5767} & \textbf{0.5315}& \textbf{0.6932} & \textbf{0.4411} & \textbf{0.4460}& \textbf{0.5447} & \textbf{0.5036} & \textbf{0.3888}\\
\bottomrule[1pt]
\end{tabu}
\vspace{-2mm}
\label{clustering_results}
\end{table*}

\section{Experiments}
\label{section4}
\subsection{Datasets}
We use four network datasets (Cora, Citeseer, Wiki, and Pubmed) and three image datasets (COIL20, YALE, and MNIST) for node clustering and link prediction tasks.
Every network dataset has the feature matrix $X$ and the affinity matrix $A$ and every image dataset has the feature matrix $X$ only.
The summary of each dataset are presented in Table \ref{dataset} and details are reported in the supplementary material.
Also, the sample images of each image dataset are described in Figure \ref{sample_image}.
\begin{table}[h]
\caption{Summary of datasets}
\vspace{-3mm}
\footnotesize
\begin{tabu}{c | X[c] | X[c] | X[c] | X[c]}
\toprule[1pt]
& $\#$ Nodes & Dimension & Classes & $\#$ Edges \\
\midrule
Cora\cite{sen2008collective} & 2708 & 1433 & 7 & 5429 \\
Citeseer\cite{sen2008collective} & 3312 & 3703 & 6 & 4732 \\
Wiki\cite{yang2015network} & 2405 & 4973 & 17 & 17981 \\
Pubmed\cite{sen2008collective} & 19717 & 500 & 3 & 44338 \\
COIL20\cite{nene1996columbia} & 1440 & 1024 & 20 & $-$ \\
YALE\cite{georghiades2001few} & 5850 & 1200 & 10 & $-$ \\
MNIST\cite{lecun1998mnist} & 10000 & 784 & 10 & $-$ \\
\bottomrule[1pt]
\end{tabu}
\vspace{-2mm}
\label{dataset}
\end{table}

\subsection{Experimental settings}
\begin{table*}[t]
\caption{Experimental results of image clustering}
\vspace{-3mm}
\footnotesize
\begin{tabu}{c | X[c] X[c] X[c] | X[c] X[c] X[c] | X[c] X[c] X[c]}
\toprule[1pt]
 \multirow{2}{*}{} & & COIL20 & & & YALE & & & MNIST & \\
 \cmidrule{2-10}
 & ACC & NMI & ARI & ACC & NMI & ARI & ACC & NMI & ARI  \\
\midrule
\midrule
 Kmeans\cite{lloyd1982least} & 0.6118 & 0.7541 & 0.5545 & 0.7450 & 0.8715 & 0.7394 & 0.5628 & 0.5450 & 0.4213 \\
 Spectral\cite{ng2002spectral} & 0.6806 & 0.8324 & 0.6190 & 0.5793 & 0.7202 & 0.4600 & 0.6496 & 0.7204 & 0.5836 \\
 GAE\cite{kipf2016variational} & 0.6632 & 0.7420 & 0.5514 & 0.8520 & 0.8851 & 0.8122 & 0.7043 & 0.6535 & 0.5534 \\
 VGAE\cite{kipf2016variational} & 0.6847 & 0.7465 & 0.5627 & 0.9157 & 0.9358 & 0.8873 & 0.7163 & 0.7149 & 0.6154 \\
 MGAE\cite{wang2017mgae} & 0.6507 & 0.7889 & 0.6004 & 0.8203 & 0.8550 & 0.7636 & 0.5807 & 0.5820 & 0.4362 \\
 ARGA\cite{pan2018adversarially} & 0.7271 & 0.7895 & 0.6183 & 0.9309 & 0.9394 & 0.8961 & 0.6672 & 0.6759 & 0.5552 \\
 ARVGA\cite{pan2018adversarially} & 0.7222 & 0.7917 & 0.6240 & 0.8727 & 0.8803 & 0.7944 & 0.6328 & 0.6123 & 0.4909 \\ 
 \midrule
 GALA & 0.8000 & 0.8771 & 0.7550 & 0.8530 & 0.9486 & 0.8647 & 0.7384 & 0.7506 & 0.6469 \\
 GALA+SCC & \textbf{0.8229} & \textbf{0.8851} & \textbf{0.7579} & \textbf{0.9933} & \textbf{0.9860} & \textbf{0.9854} & \textbf{0.7426} & \textbf{0.7565} & \textbf{0.6675} \\
\bottomrule[1pt]
\end{tabu}
\vspace{-2mm}
\label{image_results}
\end{table*}

To measure the performance of node clustering task, we use three metrics: accuracy (ACC), normalized mutual information (NMI), and adjusted rand index (ARI) as in \cite{wang2017mgae}. 
We report the mean values of the three metrics for each algorithm after executing 50 times, and the higher values imply the more correct results.
For link prediction task, we partitioned the dataset following the work of GAE \cite{kipf2016variational}, and reported mean scores and standard errors of Area Under Curve (AUC) and Average Precision (AP) with 10 random initializations.
The implementation details such as hyperparameters are reported in the supplementary material.
\begin{table}[t!]
\caption{Experiment results on Pubmed dataset}
\vspace{-1mm}
\footnotesize
\begin{tabu}{c | X[c] | X[c] | X[c]}
\toprule[1pt]
& ACC & NMI & ARI \\
\midrule
\midrule
Kmeans\cite{lloyd1982least} & 0.5952 & 0.3152 & 0.2817 \\
Spectral\cite{ng2002spectral} & 0.5282 & 0.0971 & 0.0620 \\
GAE\cite{kipf2016variational} & 0.6861 & 0.2957 & 0.3046 \\
VGAE\cite{kipf2016variational} & 0.6887 & 0.3108 & 0.3018 \\
MGAE\cite{wang2017mgae} & 0.5932 & 0.2822 & 0.2483 \\
ARGA\cite{pan2018adversarially} & 0.6807 & 0.2757 & 0.2910 \\
ARVGA\cite{pan2018adversarially} & 0.5130 & 0.1169 & 0.0777 \\
\midrule
GALA & \textbf{0.6939} & \textbf{0.3273} & \textbf{0.3214} \\
\bottomrule[1pt]
\end{tabu}
\vspace{-2mm}
\label{pubmed}
\end{table}
\subsection{Comparing methods}
We compare the performance of 15 algorithms.
Compared algorithms can be categorized into four groups as described below: 

\begin{itemize}[leftmargin=*]
\item \textbf{i) Using features only.}
`Kmeans' \cite{lloyd1982least} is the K-means clustering based on only the features of the data, which is the baseline clustering algorithm in our experiment.
\item \textbf{ii) Using network structures only.}
`Spectral' \cite{ng2002spectral} is a spectral clustering algorithm using eigendecomposition on graph Laplacian.
`Big-Clam' \cite{yang2013overlapping} is a large-scale community detection algorithm utilizing a variant of nonnegative matrix factorization.
`DeepWalk' \cite{perozzi2014deepwalk} learns the latent social representation of nodes using local information through a neural network.
`GraEnc' \cite{tian2014learning} is a graph-encoding neural network derived from the relation between autoencoder and spectral clustering.
`DNGR' \cite{cao2016deep} generates a low-dimensional representation of each node by using a graph structure and a stacked denoised autoencoder.
\item \textbf{iii) Using both.}
`Circles' \cite{leskovec2012learning} is an algorithm which discovers social circles through a node clustering algorithm.
`RTM' \cite{chang2009relational} presents a relational topic model of documents and links between the documents.
`RMSC' \cite{xia2014robust} is a robust multi-view spectral clustering algorithm which can handle noises in the data and recover transition matrix through low-rank and sparse decomposition.
`TADW' \cite{yang2015network} interprets DeepWalk from the view of matrix factorization and incorporates text features of nodes.  
\item \textbf{iv) Using both with spectral convolution on graphs.} 
`GAE' \cite{kipf2016variational} is the first attempt to graft the spectral convolution on graphs onto autoencoder framework.
`VGAE' \cite{kipf2016variational} is the variational variant of GAE.
`MGAE' \cite{wang2017mgae} is an autoencoder which combines the marginalization process with spectral convolution on graphs.
`ARGA' \cite{pan2018adversarially} learns the latent representation by adding an adversarial model to a non-probabilistic variant of VGAE.
`ARVGA' \cite{pan2018adversarially} is an algorithm which adds an adversarial model to VGAE.
\end{itemize}
\subsection{Node clustering results}
\begin{table*}[t]
\caption{Effects of stable decoder and subspace clustering cost}
\vspace{-3mm}
\footnotesize
\begin{tabu}{c | X[c] X[c] X[c] | X[c] X[c] X[c] | X[c] X[c] X[c]}
\toprule[1pt]
 \multirow{2}{*}{} & & COIL20 & & & YALE & & & MNIST & \\
 \cmidrule{2-10}
 & ACC & NMI & ARI & ACC & NMI & ARI & ACC & NMI & ARI  \\
\midrule
\midrule
 \makecell{Unstable decoder and reconstruction cost only \\ (Eq. \ref{propagation} and Eq. \ref{recon_loss})} & 0.5961 & 0.7986 & 0.5492 & 0.7205 & 0.9028 & 0.7530 & 0.6589 & 0.7397 & 0.5983 \\ 
 \makecell{Unstable decoder and subspace clustering cost \\ (Eq. \ref{propagation} and Eq. \ref{final_loss})} & 0.7104 & 0.8074 & 0.6429 & 0.7810 & 0.8710 & 0.7130 & 0.6734 & 0.7211 & 0.6028 \\
 \makecell{Stable decoder and reconstruction cost only \\ (Eq. \ref{prop_enc} and Eq. \ref{recon_loss})} & 0.8000 & 0.8771 & 0.7550 & 0.8530 & 0.9486 & 0.8646 & 0.7384 & 0.7506 & 0.6469 \\ 
 \midrule
 \makecell{Stable decoder and subspace clustering cost \\ (Eq. \ref{prop_enc} and Eq. \ref{final_loss})} & \textbf{0.8229} & \textbf{0.8851} & \textbf{0.7579}& \textbf{0.9933} & \textbf{0.9860} & \textbf{0.9854} & \textbf{0.7426} & \textbf{0.7565} & \textbf{0.6675} \\
\bottomrule[1pt]
\end{tabu}
\vspace{-2mm}
\label{Ablation}
\end{table*}

The experimental results of node clustering are presented in Table \ref{clustering_results}.
It can be observed that for every dataset, the methods which use features and network structures simultaneously show better performance than the methods which use only one of them.
Furthermore, among the methods which use both features and network structures, algorithms with neural network models which exploit spectral convolution on graphs present outstanding performance since they can learn deeper relationships between nodes than the methods which do not use spectral convolution on graphs.
In 
every experiments, GALA shows superior performance to other methods.
Especially, for the Cora dataset, GALA outperforms VGAE, which is the first graph convolution autoencoder framework, by about 
24.39$\%$, 24.75$\%$ and 27.68$\%$, and MGAE, which is the state-of-the-art graph convolutional autoencoder algorithm, by about 
6.15$\%$, 6.56$\%$ and 8.68$\%$ on ACC, NMI and ARI, respectively.
The better performance of GALA comes from the better decoder design based on the numerically stable form of Laplacian sharpening both and full utilizing of graph structure and node attributes in the whole autoencoder framework.

Furthermore, we conduct another node clustering experiment on a large network dataset (Pubmed), and the results are reported in Table \ref{pubmed}.
We can observe that GALA outperforms every baselines and state-of-the-art graph convolution algorithms.
Although Kmeans clustering, a baseline algorithm, shows higher performance over several graph convolution algorithms on NMI and ARI, the proposed method presents better performances.

\subsection{Image clustering results}
The experimental results of image clustering are presented in Table \ref{image_results}.
We report both GALA's performance of reconstruction cost only case and the subspace clustering cost added case (GALA+SCC).
It can be seen that GALA outperforms several baselines and the state-of-the-art graph convolution algorithms for most of the cases.
Also, for every case, the proposed subspace clustering cost term contributes to improve the performance of the image clustering.
On the YALE dataset, notably, we can observe that the proposed subspace clustering cost term significantly enhances the image clustering performance and achieves nearly perfect accuracy.

\begin{figure*}[h!]
\centering
\begin{subfigure}{.25\textwidth}
\centering
\includegraphics[clip, trim = 2.5cm 1.7cm 2cm 1.6cm, width=0.7\linewidth]{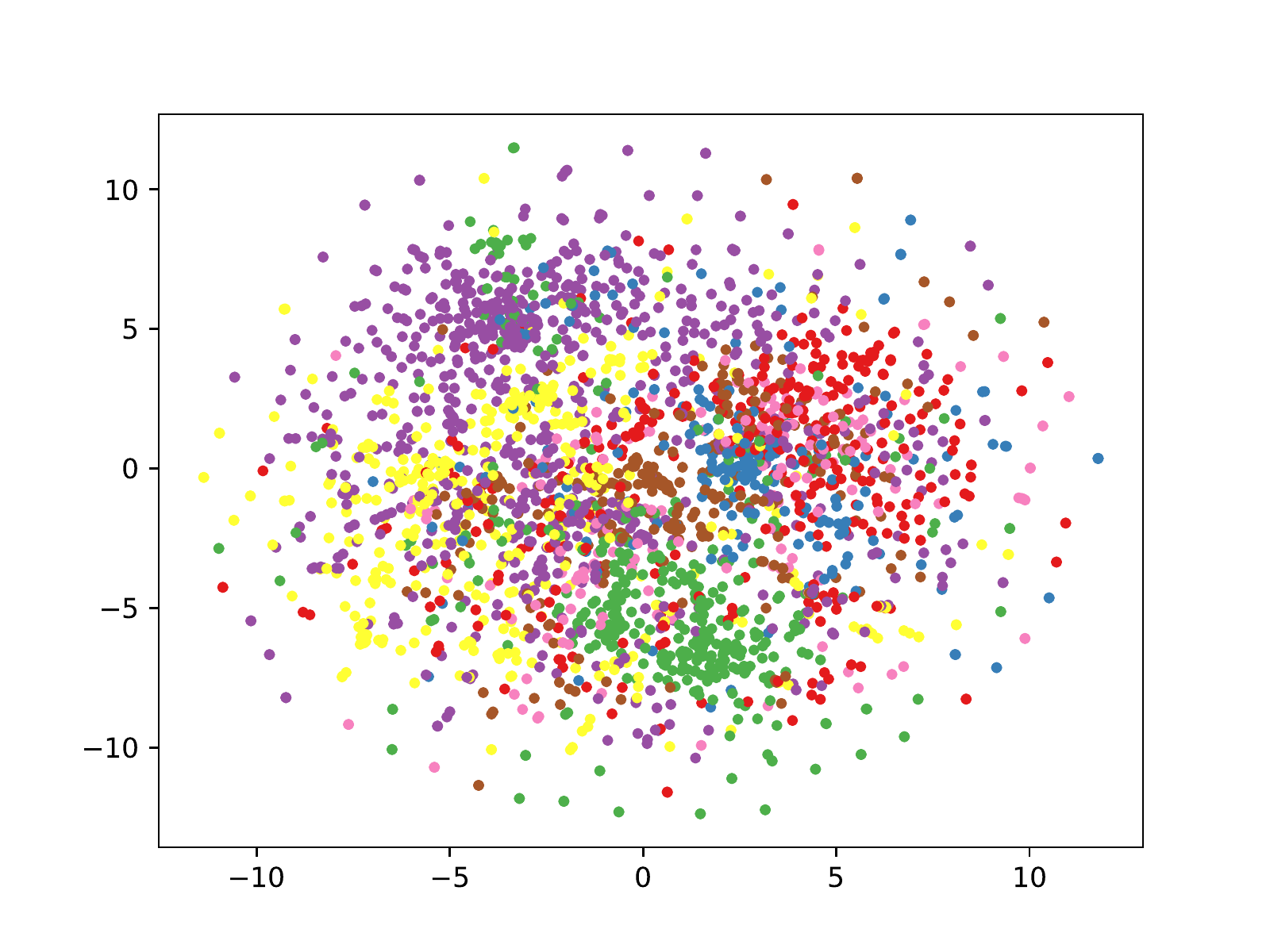}
\caption{Cora (raw) \\ \textcolor{white}{f} \hspace{0.3cm}}
\end{subfigure}\hfill
\begin{subfigure}{.25\textwidth}
\centering
\includegraphics[clip, trim = 2.5cm 1.7cm 2cm 1.6cm, width=0.7\linewidth]{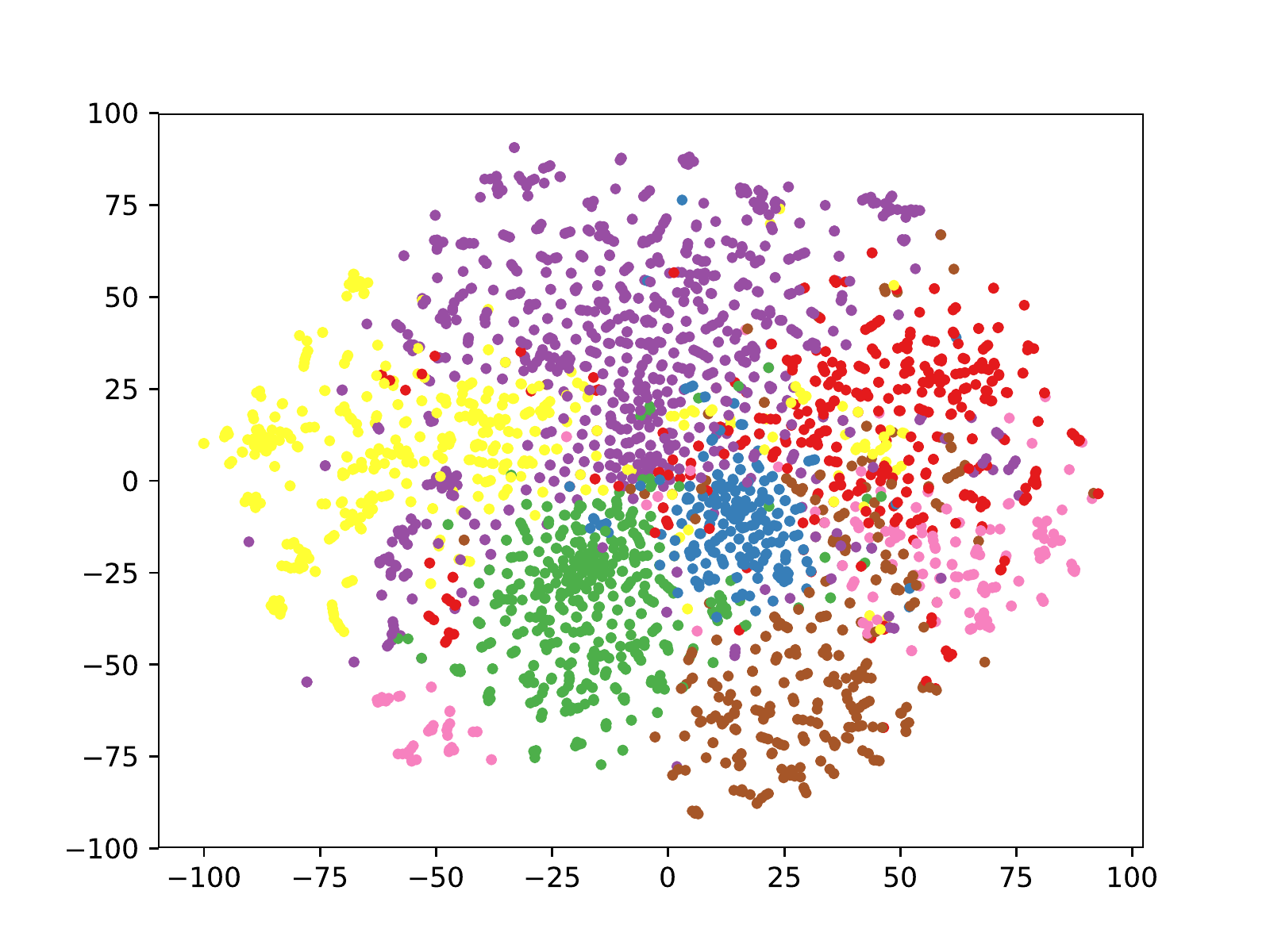}
\caption{Cora (GALA) \\ \textcolor{white}{f} \hspace{0.3cm}}
\end{subfigure}\hfill
\begin{subfigure}{.25\textwidth}
\centering
\includegraphics[clip, trim = 2.5cm 1.7cm 2cm 1.6cm, width=0.7\linewidth]{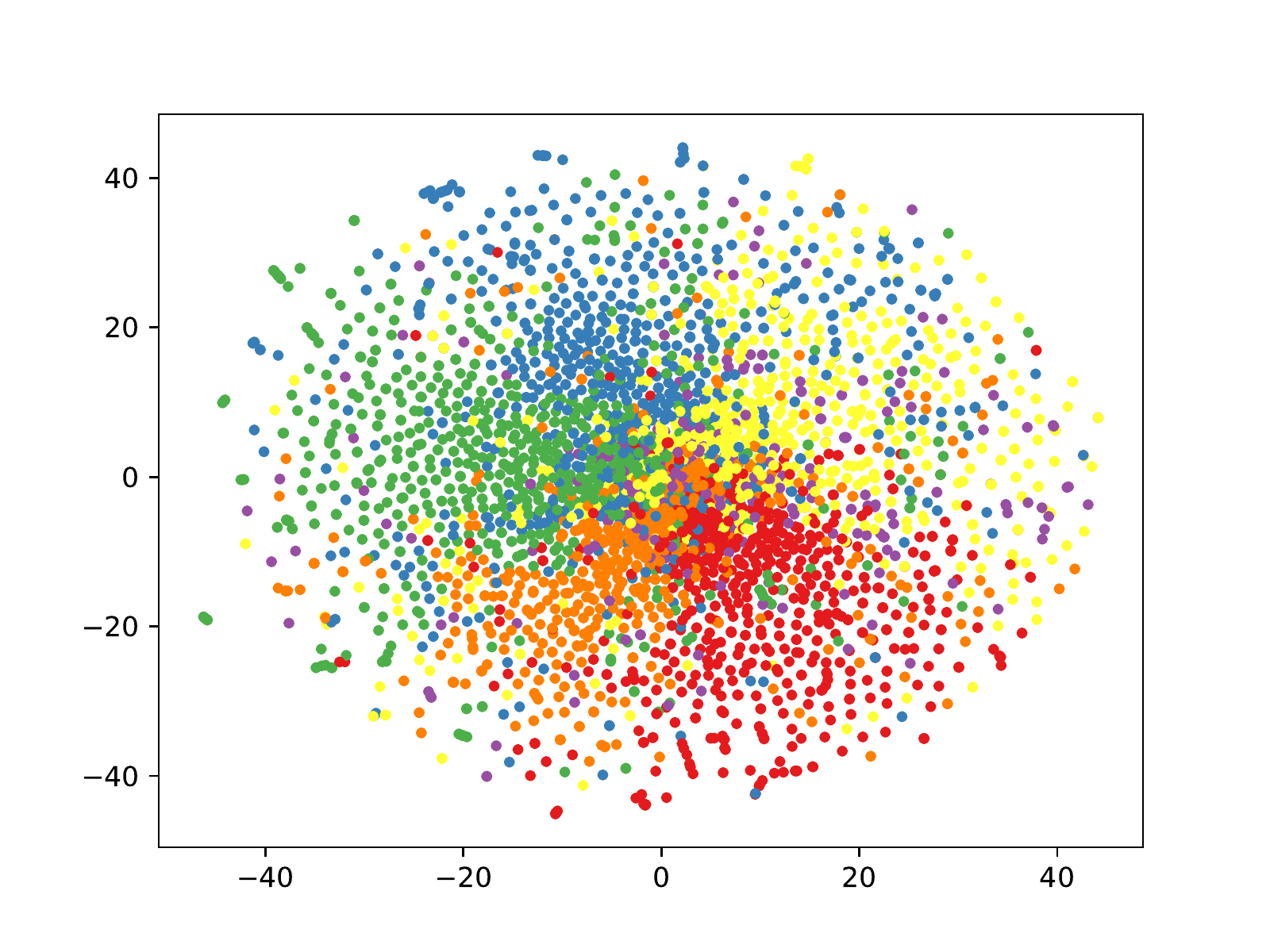}
\caption{Citeseer (raw) \\ \textcolor{white}{f} \hspace{0.3cm}}
\end{subfigure}\hfill
\begin{subfigure}{.25\textwidth}
\centering
\includegraphics[clip, trim = 2.5cm 1.7cm 2cm 1.6cm, width=0.7\linewidth]{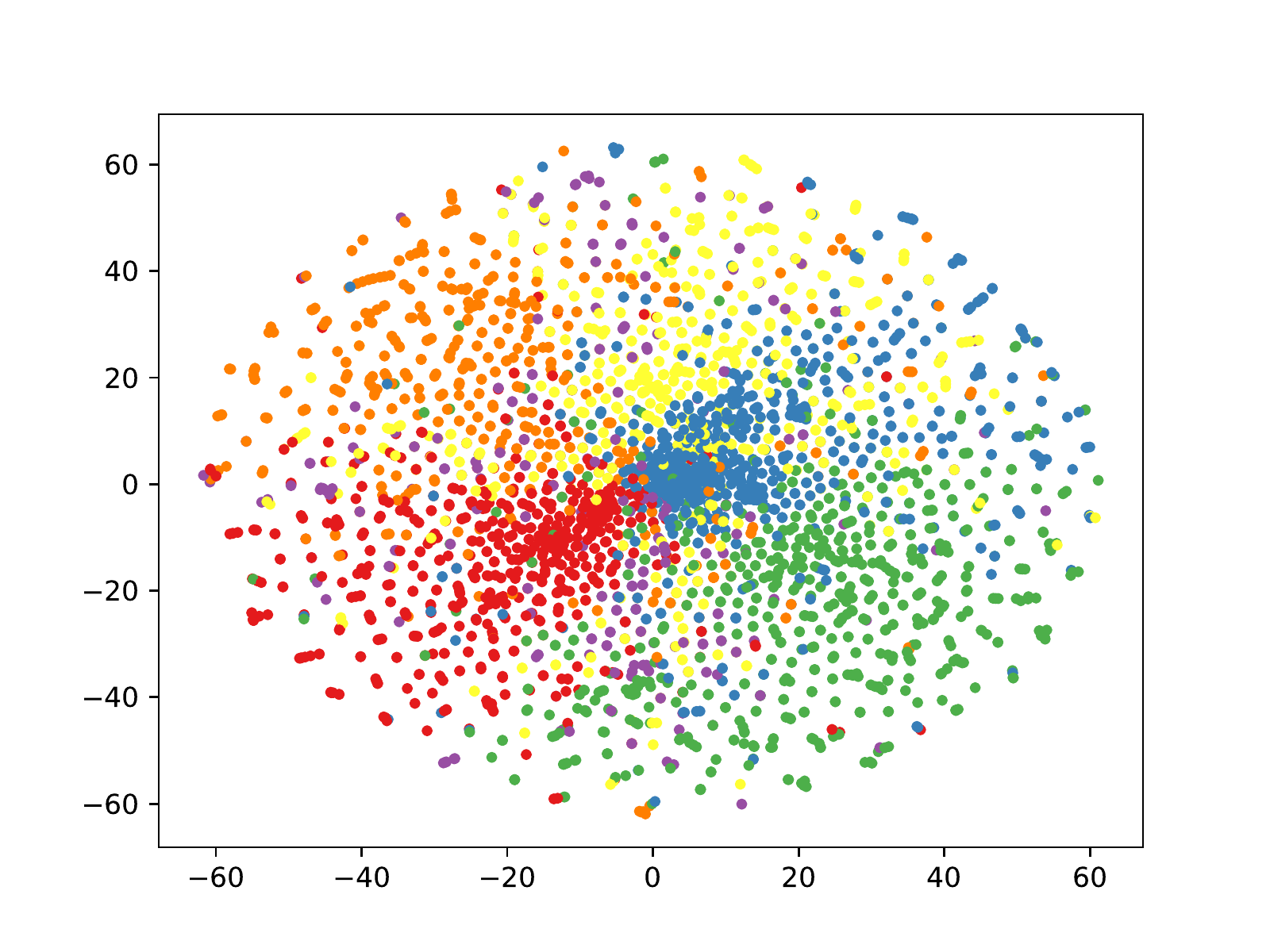}
\caption{Citeseer (GALA) \\ \textcolor{white}{f} \hspace{0.3cm}}
\end{subfigure}\par
\begin{subfigure}{.2\textwidth}
\centering
\includegraphics[clip, trim = 2.8cm 1cm 2.2cm 0.6cm, width=0.95\linewidth]{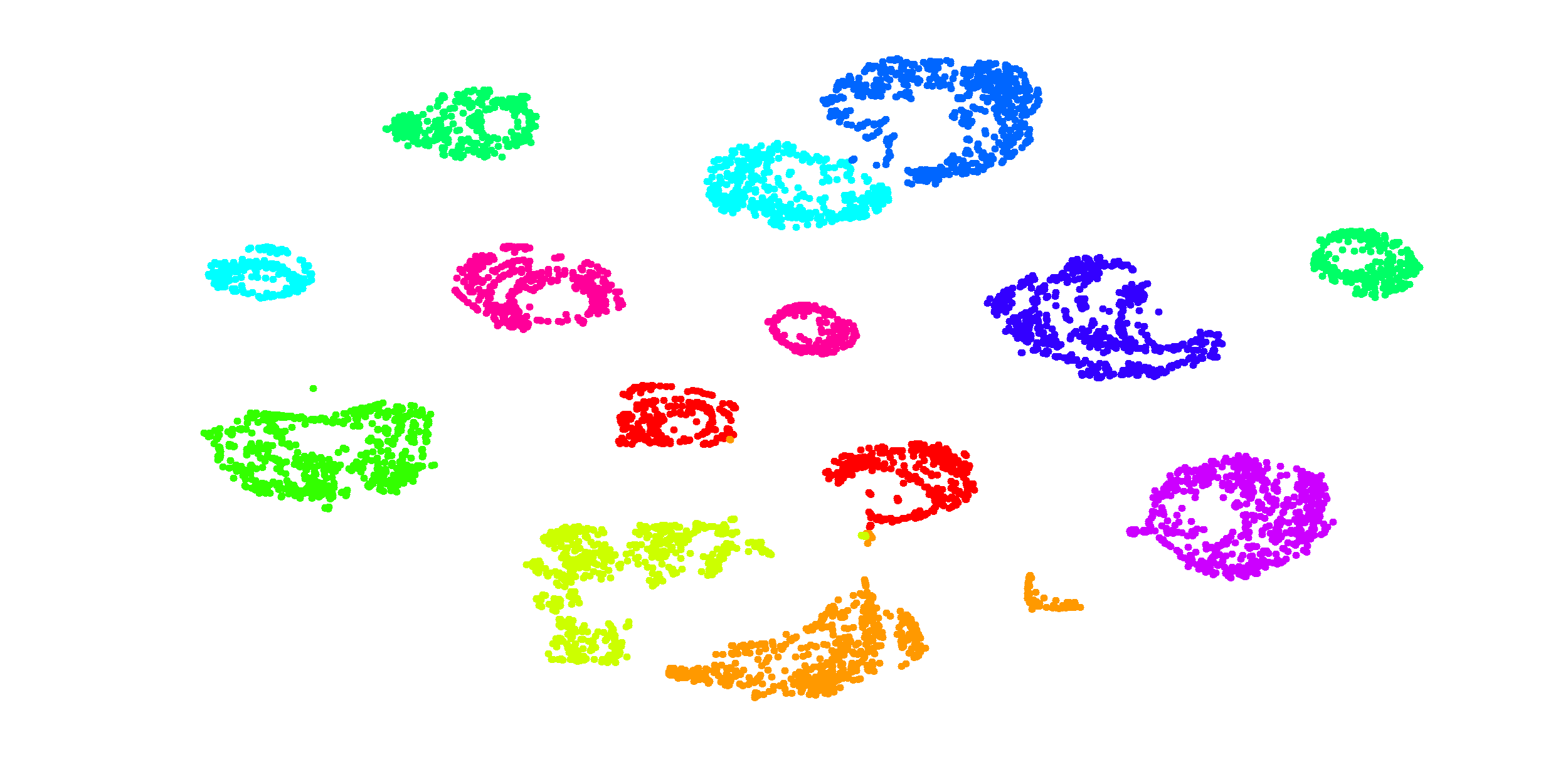}
\caption{YALE (raw) \\ \textcolor{white}{f} \hspace{0.3cm}}
\end{subfigure}
\begin{subfigure}{.195\textwidth}
\centering
\includegraphics[clip, trim =  2.8cm 1cm 2.2cm 0.6cm, width=0.95\linewidth]{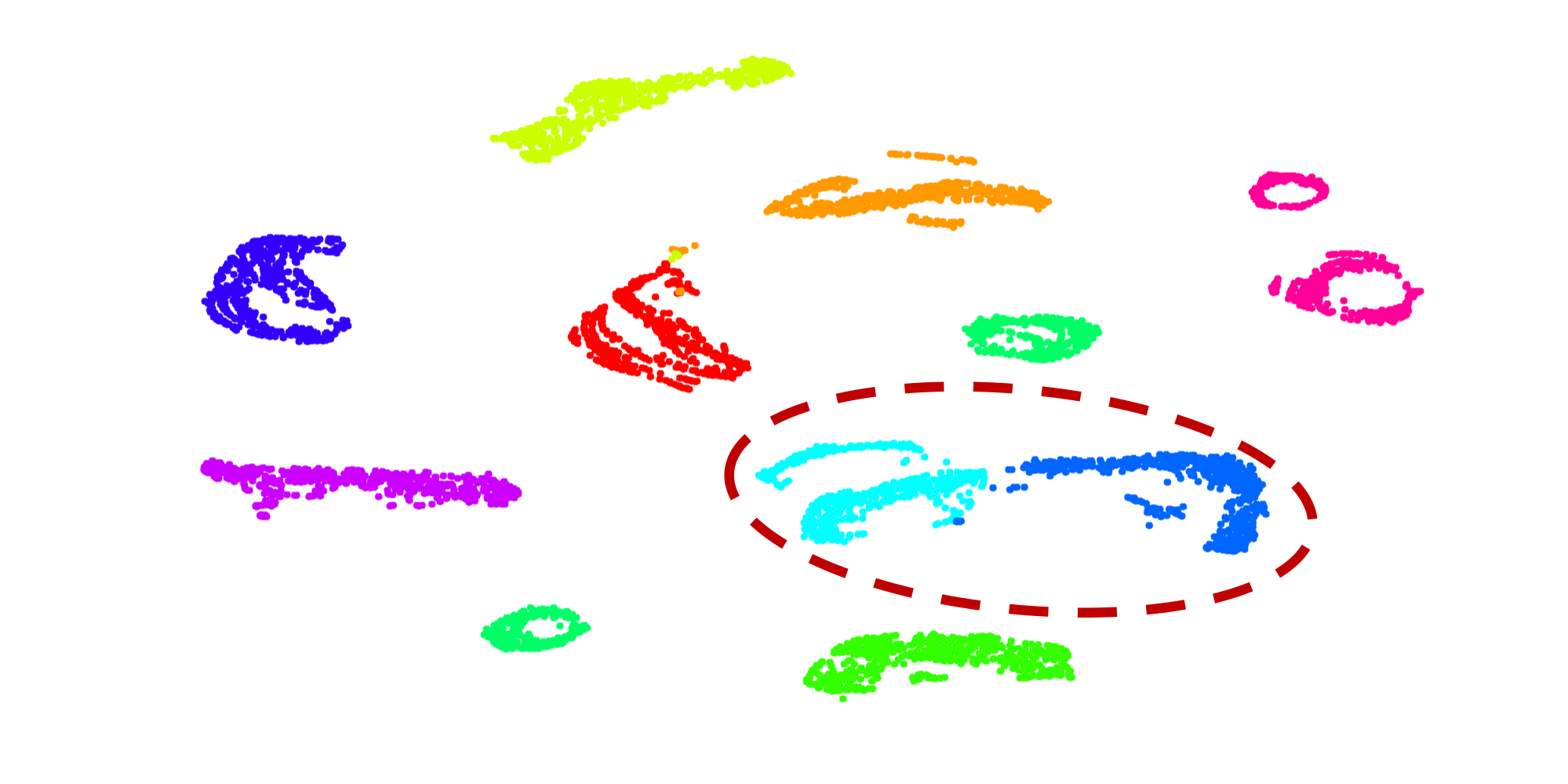}
\caption{YALE (VGAE) \\ \textcolor{white}{f} \hspace{0.3cm}}
\end{subfigure}
\begin{subfigure}{.195\textwidth}
\centering
\includegraphics[clip, trim =  2.8cm 1cm 2.2cm 0.6cm, width=0.95\linewidth]{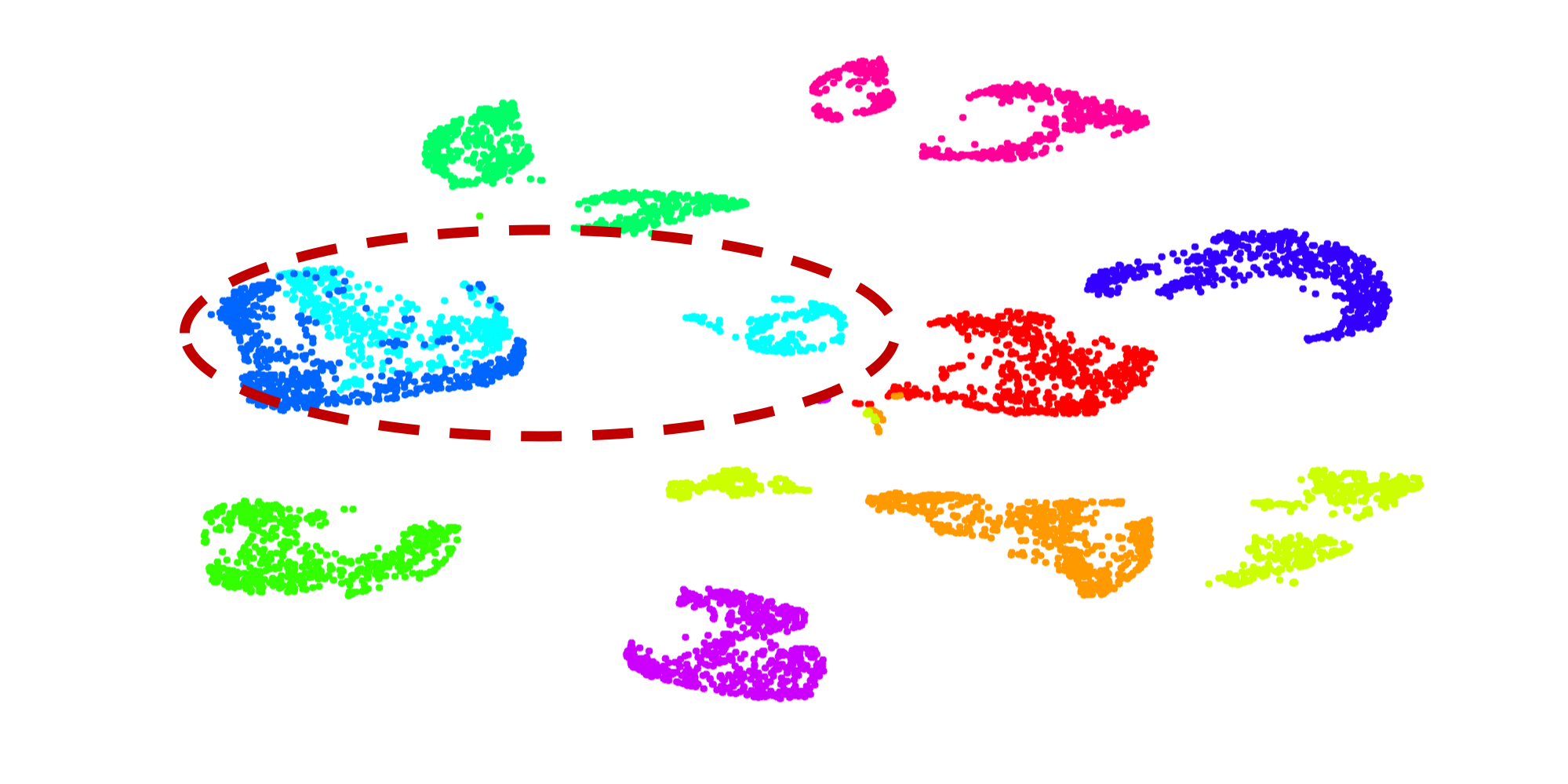}
\caption{YALE (MGAE) \\ \textcolor{white}{f} \hspace{0.3cm}}
\end{subfigure}\hfill
\begin{subfigure}{.195\textwidth}
\centering
\includegraphics[clip, trim =  2.8cm 1cm 2.2cm 0.6cm, width=0.95\linewidth]{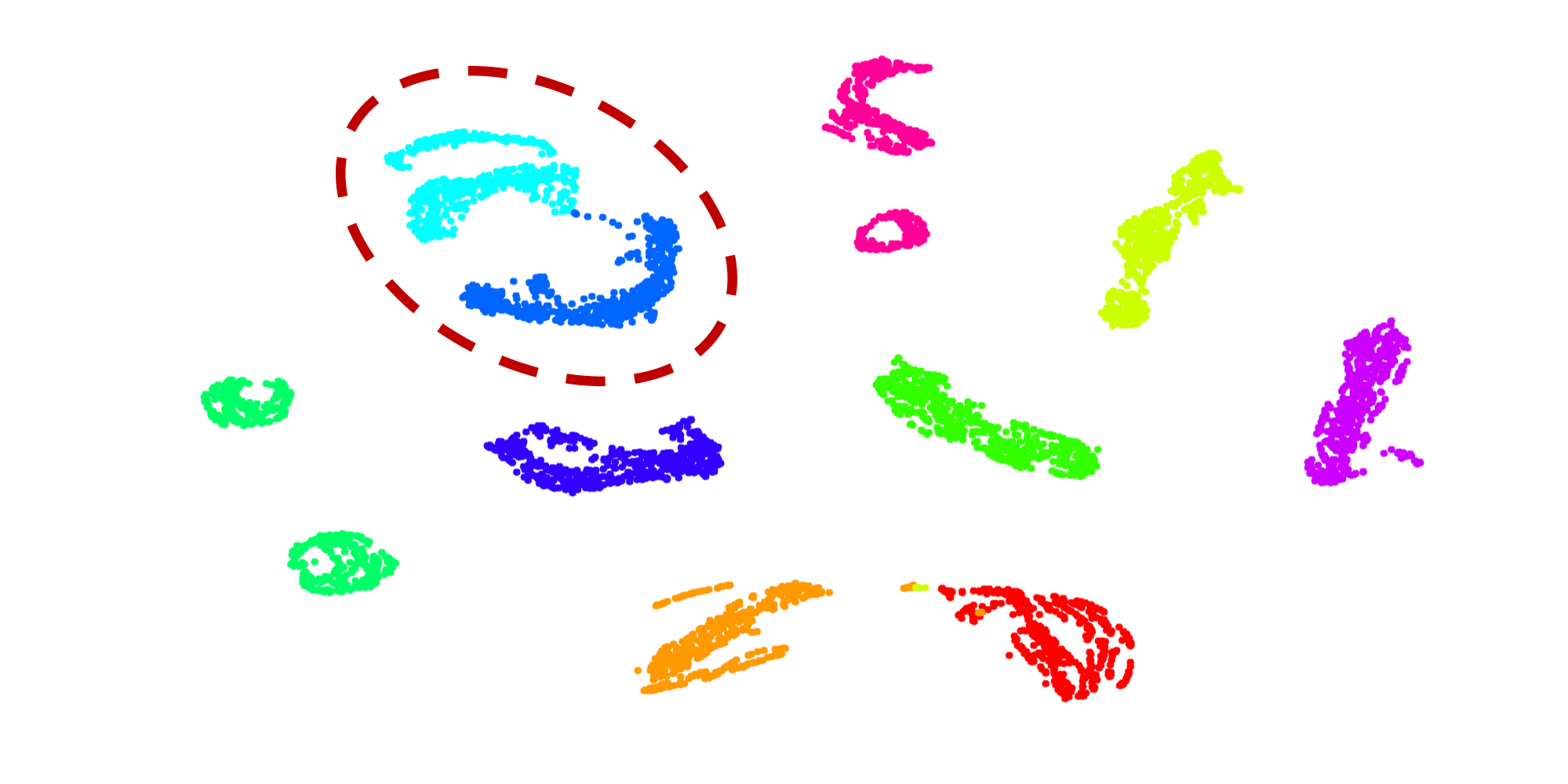}
\caption{YALE (ARGA) \\ \textcolor{white}{f} \hspace{0.3cm}}
\end{subfigure}\hfill
\begin{subfigure}{.195\textwidth}
\centering
\includegraphics[clip, trim =  2.8cm 1cm 2.2cm 0.6cm, width=0.95\linewidth]{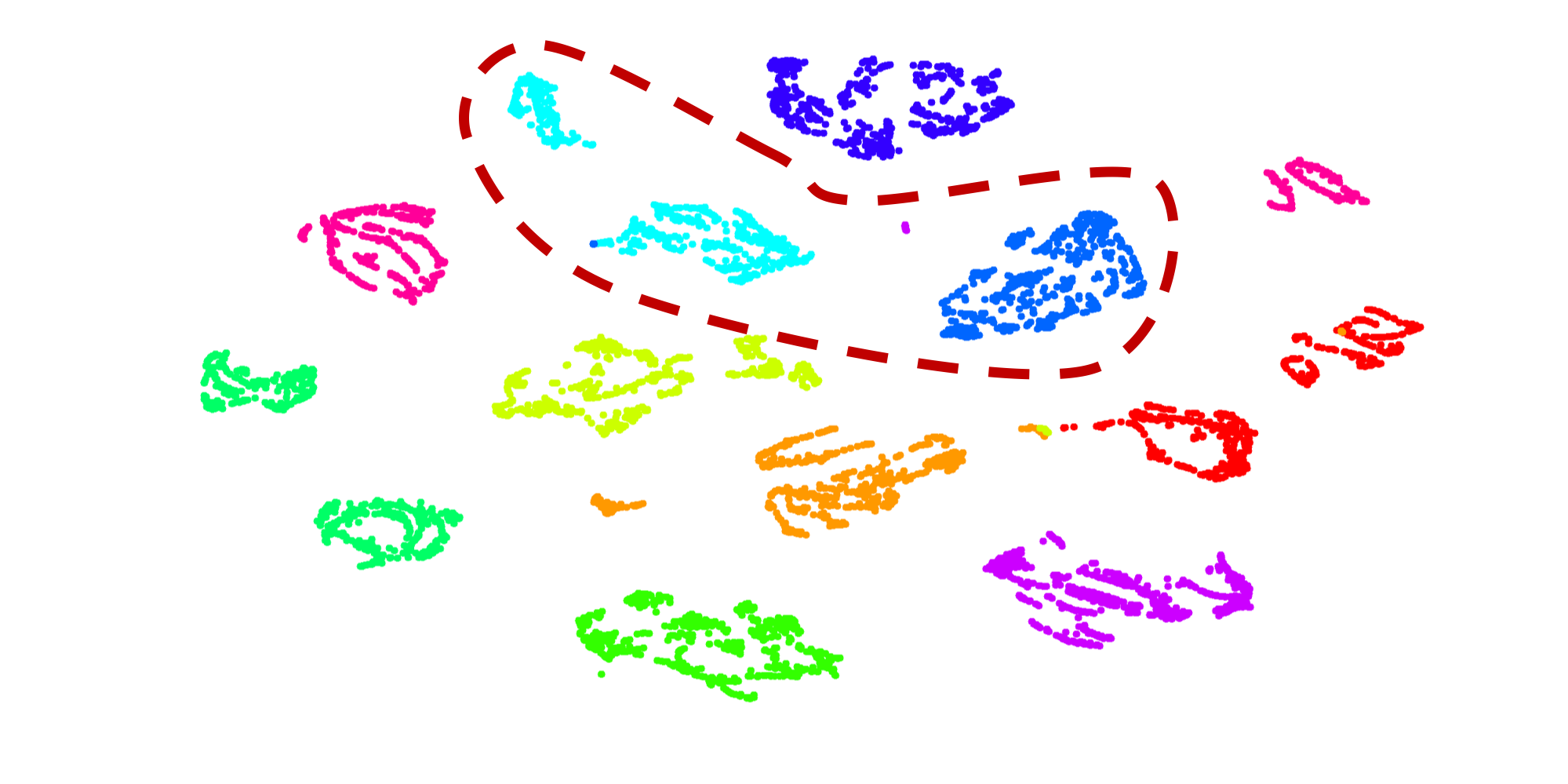}
\caption{YALE (GALA) \\ \textcolor{white}{f} \hspace{0.3cm}}
\end{subfigure}\hfill
\vspace{-6mm}
\caption{
The two-dimensional visualizations of raw features of each node and the latent representations of compared methods and GALA for Cora, Citeseer and YALE are presented.
The same color indicates the same cluster. 
}
\label{visualization}
\vspace{-2mm}
\end{figure*}
\subsection{Ablation studies}
We validate the effectiveness of the proposed stable decoder and the subspace clustering cost by image clustering experiments on the three image datasets (COIL20, YALE and MNIST). 
There are four configurations as shown in Table \ref{Ablation}. We would like to note that the reconstruction cost only (Eq. \ref{recon_loss}) is a subset of subspace clustering cost (Eq. \ref{final_loss}), thus the last configuration is the full proposed method.
Reported numbers are mean values after executing 50 times.
It can be clearly noticed that the numerically stable form of Laplacian sharpening and subspace clustering cost are helpful to find the latent representations which reflect the graph structures certainly and using both components can boost the performance of clustering.
In addition, it can be seen that the stable decoder with the reconstruction cost only outperforms the state-of-the-art algorithms in most cases because GALA can utilize the graph structure in the whole processes of the autoencoder architecture.


\subsection{Link prediction results}
We provide some results on link prediction task on Citeseer dataset. 
For link prediction task, we minimized the below cost function that added link prediction cost of GAE \cite{kipf2016variational} to the reconstruction cost, where $H$ is the latent representation, $\hat{A}=$ sigmoid$(HH^T)$ is the reconstructed affinity matrix and $\gamma$ is the regularization parameter.
\begin{equation}
\begin{aligned}
& \underset{\bar{X}, H}{\text{min}}
& & \frac{1}{2}{\| X - \bar{X} \|}^2_F + \gamma \mathbb{E}_{H}[\log p(\hat{A}|H)].
\end{aligned}
\label{link_loss}
\end{equation}

The results are shown in Table \ref{link_citeseer}, and our model outperforms the compared methods in terms of the link prediction task as well as the node clustering task.
\begin{table}[t]
\caption{Experimental results of link prediction on Citeseer}
\vspace{-2mm}
\footnotesize
\begin{tabu}{c | X[c] X[c]}
\toprule[1pt]
 & AUC & AP \\
\midrule
\midrule
 GAE\cite{kipf2016variational} & 89.5 $\pm$ 0.04 & 89.9 $\pm$ 0.05 \\
 VGAE\cite{kipf2016variational} & 90.8 $\pm$ 0.02 & 92.0 $\pm$ 0.02 \\
 ARGA\cite{pan2018adversarially} & 91.9 $\pm$ 0.003 & 93.0 $\pm$ 0.003 \\
 ARVGA\cite{pan2018adversarially} & 92.4 $\pm$ 0.003 & 93.0 $\pm$ 0.003 \\
 \midrule
 GALA & \textbf{94.4 $\pm$ 0.009} & \textbf{94.8 $\pm$ 0.010} \\
\bottomrule[1pt]
\end{tabu}
\label{link_citeseer}
\end{table}
\subsection{Visualization}
One of the key ideas of the proposed autoencoder is that the encoder makes the feature of each node becomes similar with its neighbors, and the decoder makes the features of each node distinguishable with its neighbors using the geometrical structure of the graphs.
To validate the proposed model, we visualize the distribution of learned latent representations and the input features of each node in two-dimensional space using \textit{t-SNE} \cite{maaten2008visualizing} as shown in Figure \ref{visualization}.
From the visualization, we can see that GALA is well-clustering the data according to their corresponding labels even though GALA performs in an unsupervised manner.
Also, we can see through the red dotted line in embedding results of the latent representation on YALE that GALA embeds the representation of nodes better than the compared methods by minimizing inter-cluster affinity and maximizing intra-cluster affinity.
\section{Conclusions}
\label{section5}
In this paper, we proposed a novel autoencoder framework which can extract low-dimensional latent representations from a graph in irregular domains.
We designed a symmetric graph convolutional autoencoder architecture where the encoder performs Laplacian smoothing while the decoder performs Laplacian sharpening.
Also, to prevent numerical instabilities, we designed a new representation of Laplacian sharpening with spectral radius one by incorporating the concept of the signed graph.
To enhance the performance of image clustering tasks, we added a subspace clustering cost term to the reconstruction cost of the autoencoder.
Experimental results on the network and image datasets demonstrated the validity of the proposed framework and had shown superior performance over various graph-based clustering algorithms.

\vspace{1mm}
\small
\noindent\textbf{Acknowledgement:} 
This work was supported by Next Generation ICD Program through NRF funded by Ministry of S\&ICT [2017M3C4A7077582], ICT R\&D program of MSIP/IITP [No.B0101-15-0552, Predictive Visual Intelligence Technology], and the Basic Science Research Program through the National Research Foundation of Korea funded by the Ministry of Science and ICT under Grant NRF-2017R1C1B2012277.

{\small
\bibliographystyle{ieee_fullname}
\bibliography{egbib}

\begin{thebibliography}{10}\itemsep=-1pt

\bibitem{belkin2002laplacian}
Mikhail Belkin and Partha Niyogi.
\newblock Laplacian eigenmaps and spectral techniques for embedding and
  clustering.
\newblock In {\em Advances in Neural Information Processing Systems}, pages
  585--591, 2002.

\bibitem{bronstein2017geometric}
Michael~M Bronstein, Joan Bruna, Yann LeCun, Arthur Szlam, and Pierre
  Vandergheynst.
\newblock Geometric deep learning: going beyond euclidean data.
\newblock {\em IEEE Signal Processing Magazine}, 34(4):18--42, 2017.

\bibitem{cao2016deep}
Shaosheng Cao, Wei Lu, and Qiongkai Xu.
\newblock Deep neural networks for learning graph representations.
\newblock In {\em AAAI}, pages 1145--1152, 2016.

\bibitem{chang2009relational}
Jonathan Chang and David Blei.
\newblock Relational topic models for document networks.
\newblock In {\em Artificial Intelligence and Statistics}, pages 81--88, 2009.

\bibitem{chung1997spectral}
Fan~RK Chung and Fan~Chung Graham.
\newblock {\em Spectral graph theory}.
\newblock Number~92. American Mathematical Soc., 1997.

\bibitem{defferrard2016convolutional}
Micha{\"e}l Defferrard, Xavier Bresson, and Pierre Vandergheynst.
\newblock Convolutional neural networks on graphs with fast localized spectral
  filtering.
\newblock In {\em Advances in Neural Information Processing Systems}, pages
  3844--3852, 2016.

\bibitem{duvenaud2015convolutional}
David~K Duvenaud, Dougal Maclaurin, Jorge Iparraguirre, Rafael Bombarell,
  Timothy Hirzel, Al{\'a}n Aspuru-Guzik, and Ryan~P Adams.
\newblock Convolutional networks on graphs for learning molecular fingerprints.
\newblock In {\em Advances in Neural Information Processing Systems}, pages
  2224--2232, 2015.

\bibitem{georghiades2001few}
Athinodoros~S Georghiades, Peter~N Belhumeur, and David~J Kriegman.
\newblock From few to many: Illumination cone models for face recognition under
  variable lighting and pose.
\newblock {\em IEEE Transactions on Pattern Analysis and Machine Intelligence},
  (6):643--660, 2001.

\bibitem{grigorchuk1999asymptotic}
Rostislav~I Grigorchuk and Andrzej Zuk.
\newblock On the asymptotic spectrum of random walks on infinite families of
  graphs.
\newblock {\em Random Walks and Discrete Potential Theory (Cortona, 1997),
  Sympos. Math}, 39:188--204, 1999.

\bibitem{hammond2011wavelets}
David~K Hammond, Pierre Vandergheynst, and R{\'e}mi Gribonval.
\newblock Wavelets on graphs via spectral graph theory.
\newblock {\em Applied and Computational Harmonic Analysis}, 30(2):129--150,
  2011.

\bibitem{ji2017deep}
Pan Ji, Tong Zhang, Hongdong Li, Mathieu Salzmann, and Ian Reid.
\newblock Deep subspace clustering networks.
\newblock In {\em Advances in Neural Information Processing Systems}, pages
  24--33, 2017.

\bibitem{kingma2014adam}
Diederik~P Kingma and Jimmy Ba.
\newblock Adam: A method for stochastic optimization.
\newblock {\em arXiv preprint arXiv:1412.6980}, 2014.

\bibitem{kipf2016variational}
Thomas~N Kipf and Max Welling.
\newblock Variational graph auto-encoders.
\newblock {\em NIPS Workshop on Bayesian Deep Learning}, 2016.

\bibitem{kipf2017semi}
Thomas~N. Kipf and Max Welling.
\newblock Semi-supervised classification with graph convolutional networks.
\newblock In {\em International Conference on Learning Representations}, 2017.

\bibitem{lazer2009life}
David Lazer, Alex~Sandy Pentland, Lada Adamic, Sinan Aral, Albert~Laszlo
  Barabasi, Devon Brewer, Nicholas Christakis, Noshir Contractor, James Fowler,
  Myron Gutmann, et~al.
\newblock Life in the network: the coming age of computational social science.
\newblock {\em Science (New York, NY)}, 323(5915):721, 2009.

\bibitem{lecun1998mnist}
Yann LeCun.
\newblock The mnist database of handwritten digits.
\newblock {\em http://yann. lecun. com/exdb/mnist/}, 1998.

\bibitem{leskovec2012learning}
Jure Leskovec and Julian~J Mcauley.
\newblock Learning to discover social circles in ego networks.
\newblock In {\em Advances in Neural Information Processing Systems}, pages
  539--547, 2012.

\bibitem{li2009note}
Hong~Hai Li and Jiong~Sheng Li.
\newblock Note on the normalized laplacian eigenvalues of signed graphs.
\newblock {\em Australas. J. Combin}, 44:153--162, 2009.

\bibitem{li2018deeper}
Qimai Li, Zhichao Han, and Xiao-Ming Wu.
\newblock Deeper insights into graph convolutional networks for semi-supervised
  learning.
\newblock In {\em Thirty-Second AAAI Conference on Artificial Intelligence},
  2018.

\bibitem{litany2017deformable}
Or Litany, Alex Bronstein, Michael Bronstein, and Ameesh Makadia.
\newblock Deformable shape completion with graph convolutional autoencoders.
\newblock {\em arXiv preprint arXiv:1712.00268}, 2017.

\bibitem{lloyd1982least}
Stuart Lloyd.
\newblock Least squares quantization in pcm.
\newblock {\em IEEE Transactions on Information Theory}, 28(2):129--137, 1982.

\bibitem{lu2012robust}
Can-Yi Lu, Hai Min, Zhong-Qiu Zhao, Lin Zhu, De-Shuang Huang, and Shuicheng
  Yan.
\newblock Robust and efficient subspace segmentation via least squares
  regression.
\newblock In {\em European conference on computer vision}, pages 347--360.
  Springer, 2012.

\bibitem{maaten2008visualizing}
Laurens van~der Maaten and Geoffrey Hinton.
\newblock Visualizing data using t-sne.
\newblock {\em Journal of machine learning research}, 9(Nov):2579--2605, 2008.

\bibitem{monti2017geometric}
Federico Monti, Michael Bronstein, and Xavier Bresson.
\newblock Geometric matrix completion with recurrent multi-graph neural
  networks.
\newblock In {\em Advances in Neural Information Processing Systems}, pages
  3697--3707, 2017.

\bibitem{nene1996columbia}
Sameer~A Nene, Shree~K Nayar, Hiroshi Murase, et~al.
\newblock Columbia object image library (coil-20).
\newblock 1996.

\bibitem{ng2002spectral}
Andrew~Y Ng, Michael~I Jordan, and Yair Weiss.
\newblock On spectral clustering: Analysis and an algorithm.
\newblock In {\em Advances in Neural Information Processing Systems}, pages
  849--856, 2002.

\bibitem{pan2018adversarially}
Shirui Pan, Ruiqi Hu, Guodong Long, Jing Jiang, Lina Yao, and Chengqi Zhang.
\newblock Adversarially regularized graph autoencoder for graph embedding.
\newblock In {\em IJCAI}, pages 2609--2615, 2018.

\bibitem{perozzi2014deepwalk}
Bryan Perozzi, Rami Al-Rfou, and Steven Skiena.
\newblock Deepwalk: Online learning of social representations.
\newblock In {\em Proceedings of the 20th ACM SIGKDD International Conference
  on Knowledge Discovery and Data mining}, pages 701--710. ACM, 2014.

\bibitem{sen2008collective}
Prithviraj Sen, Galileo Namata, Mustafa Bilgic, Lise Getoor, Brian Galligher,
  and Tina Eliassi-Rad.
\newblock Collective classification in network data.
\newblock {\em AI Magazine}, 29(3):93, 2008.

\bibitem{shi2000normalized}
Jianbo Shi and Jitendra Malik.
\newblock Normalized cuts and image segmentation.
\newblock {\em IEEE Transactions on Pattern Analysis and Machine Intelligence},
  22(8):888--905, 2000.

\bibitem{shuman2013emerging}
David~I Shuman, Sunil~K Narang, Pascal Frossard, Antonio Ortega, and Pierre
  Vandergheynst.
\newblock The emerging field of signal processing on graphs: Extending
  high-dimensional data analysis to networks and other irregular domains.
\newblock {\em IEEE Signal Processing Magazine}, 30(3):83--98, 2013.

\bibitem{taubin1995signal}
Gabriel Taubin.
\newblock A signal processing approach to fair surface design.
\newblock In {\em Proceedings of the 22nd Annual Conference on Computer
  graphics and Interactive techniques}, pages 351--358. ACM, 1995.

\bibitem{tian2014learning}
Fei Tian, Bin Gao, Qing Cui, Enhong Chen, and Tie-Yan Liu.
\newblock Learning deep representations for graph clustering.
\newblock In {\em AAAI}, pages 1293--1299, 2014.

\bibitem{vidal2011subspace}
Ren{\'e} Vidal.
\newblock Subspace clustering.
\newblock {\em IEEE Signal Processing Magazine}, 28(2):52--68, 2011.

\bibitem{wang2017mgae}
Chun Wang, Shirui Pan, Guodong Long, Xingquan Zhu, and Jing Jiang.
\newblock Mgae: Marginalized graph autoencoder for graph clustering.
\newblock In {\em Proceedings of the 2017 ACM on Conference on Information and
  Knowledge Management}, pages 889--898. ACM, 2017.

\bibitem{xia2014robust}
Rongkai Xia, Yan Pan, Lei Du, and Jian Yin.
\newblock Robust multi-view spectral clustering via low-rank and sparse
  decomposition.
\newblock In {\em AAAI}, pages 2149--2155, 2014.

\bibitem{yang2015network}
Cheng Yang, Zhiyuan Liu, Deli Zhao, Maosong Sun, and Edward~Y Chang.
\newblock Network representation learning with rich text information.
\newblock In {\em IJCAI}, pages 2111--2117, 2015.

\bibitem{yang2013overlapping}
Jaewon Yang and Jure Leskovec.
\newblock Overlapping community detection at scale: a nonnegative matrix
  factorization approach.
\newblock In {\em Proceedings of the sixth ACM International Conference on Web
  Search and Data Mining}, pages 587--596. ACM, 2013.

\bibitem{zhou2018deep}
Pan Zhou, Yunqing Hou, and Jiashi Feng.
\newblock Deep adversarial subspace clustering.
\newblock In {\em Proceedings of the IEEE Conference on Computer Vision and
  Pattern Recognition}, pages 1596--1604, 2018.

\end{thebibliography}
}

\end{document}